# A Bayesian Matrix Factorization Model for Relational Data

**Ajit P. Singh**
Department of Electrical Engineering
University of Washington
Seattle, WA 98195, USA
ajit@ee.washington.edu

**Geoffrey J. Gordon**
Machine Learning Department
Carnegie Mellon University
Pittsburgh, PA 15213, USA
ggordon@cs.cmu.edu

## Abstract

Relational learning can be used to augment one data source with other correlated sources of information, to improve predictive accuracy. We frame a large class of relational learning problems as matrix factorization problems, and propose a hierarchical Bayesian model. Training our Bayesian model using random-walk Metropolis-Hastings is impractically slow, and so we develop a block Metropolis-Hastings sampler which uses the gradient and Hessian of the likelihood to dynamically tune the proposal. We demonstrate that a predictive model of brain response to stimuli can be improved by augmenting it with side information about the stimuli.

## 1 Introduction

In attribute-value learning, entities are represented by a predetermined set of features, or *attributes*. Entities are then represented as records, or tuples of assignments to attributes, and are assumed to be fully described by their attributes (i.e., records are exchangeable). In relational learning, by contrast, objects have both attributes and *relations*, properties involving multiple attributes. We take a broad view of relations, as functional mappings from tuples of entities to a subset of the real numbers. Observing relational information correlates entities, and so we seek to exploit these correlations to improve predictive accuracy.[1]

**Target application for relational learning**: Functional Magnetic Resonance Imaging (fMRI) is often used to measure responses in small regions of the brain (i.e., voxels) to external stimuli. In this paper, we consider stimuli which are word-picture pairs displayed on a screen. Given enough experiments, on a sufficiently broad range of stimuli, one can build models that predict patterns of brain activation given new stimuli [9]. The fMRI data can be viewed as a relation, `Response`$(stimulus, voxel) \in [0, 1]$, measuring the response in a region of the brain under a particular stimulus, averaged over all patients. Running fMRI experiments is costly, but we can often collect cheap side information about the stimuli: e.g., we can collect statistics of whether the stimulus word co-occurs with other commonly-used words in large, freely available text corpora. The stimulus side-information can be viewed as a relation `Co-occurs`$(word, stimulus) \in \{0, 1\}$, measuring whether or not a word co-occurs near a stimulus word in the corpus. Both relations provide information about the same stimuli, and we seek to improve the quality of a predictive model of brain activity, the `Response` relation, using word co-occurrence data, the `Co-occurs` relation.

**Overview**: Our goal is to predict unobserved values of the `Response` relation. We begin with a baseline model that represents each relation as a matrix, and jointly factors the matrices under regularized maximum likelihood (Section 2). The performance of the baseline model is poor. We first attempt to address the limitations of the baseline model by introducing a hierarchical prior, using maximum a posteriori (MAP) inference for computational efficiency (Section 3). Unfortunately, the performance of the hierarchical MAP model is not significantly better than the baseline. We attribute this failure to the poor match between MAP inference and relational learning: typical arguments of asymptotic consistency do not apply when the number of parameters to estimate grows as quickly as the available data. This observation leads us to seek the full posterior distribution for our same hierarchical model (Section 4); and indeed, when we do so, performance vastly improves on our target application. Modeling the full posterior distribution is computationally expensive: the naïve block Metropolis-Hastings approach, using the common random-walk proposal, is impractically slow to mix in our case. We exploit the fact that matrix factorization models are bilinear to reduce the fully Bayesian inference problem to training a set of tied Bayesian Generalized Linear Models (GLMs). This reduction allows us to use the local gradient and Hessian of the likelihood to create an adaptive proposal distribution (Section 4.2). The adaptive proposal distributions eliminate the need for tedious hand tuning of a large number of proposal distributions, making block Metropolis-Hastings practical.

[1] An extended version of this paper appears in Singh [18].

## 2 Collective Matrix Factorization

An arity-two relation can be represented as a matrix, and sets of correlated relations can be represented as sets of matrices which share dimensions. In our fMRI example, the `Co-occurs` relation is an $m \times n$ matrix $X$; the `Response` relation is an $n \times r$ matrix $Y$. We embed the entities in a $k$-dimensional space, by factoring an indirect representation of each matrix, $X \approx f(UV^T)$ and $Y \approx g(VZ^T)$. Thus, $U$ is an $m \times k$ factor representing words, $Z$ is an $r \times k$ factor representing voxels, and $V$ is an $n \times k$ factor representing stimuli. $f$ and $g$ are link functions. Given parameters $\mathcal{F} = (U, V, Z)$, and data $\mathcal{D} = (X, Y)$, the likelihood of each matrix is

$$p(X \mid U, V, W) = \prod_{i=1}^{m} \prod_{j=1}^{n} \left[ p_X \left( X_{ij} \mid U_{i\cdot} V_{j\cdot}^T \right) \right]^{W_{ij}}, \quad (1)$$

$$p(Y \mid V, Z, \tilde{W}) = \prod_{j=1}^{n} \prod_{p=1}^{r} \left[ p_Y \left( Y_{jr} \mid V_{j\cdot} Z_{p\cdot}^T \right) \right]^{\tilde{W}_{jp}}. \quad (2)$$

The per-entry distributions $p_X$ and $p_Y$ are one-parameter exponential families with natural parameters $U_{i\cdot} V_{j\cdot}^T$ and $V_{j\cdot} Z_{p\cdot}^T$, respectively. The modeler chooses $p_X$ and $p_Y$, and they need not be from the same exponential family. This allows us to integrate relations with different response types: e.g., `Co-occurs` is well-modelled by the Bernoulli distribution, but `Response` is better modelled by a Gaussian. The fixed weights $W_{ij} \in \{0,1\}$ and $\tilde{W}_{jp} \in \{0,1\}$ allow for missing data: set a weight to zero when the corresponding value in the data matrix is unobserved. Maximizing Equation 1 is a weighted version of Exponential Family PCA [4].

Maximizing the product of Equations 1 and 2 with respect to the factors $\mathcal{F}$ is an example of collective matrix factorization [19]. Given that the number of parameters grows with the data, we place a multivariate Gaussian prior on each row of U:

$$p(U \mid \Theta_U) = \prod_{i=1}^{m} \mathcal{N}(U_{i\cdot} \mid \mu_U, \Sigma_U), \quad (3)$$

where $\mathcal{N}(\cdot \mid \mu_U, \Sigma_U)$ is a Gaussian with mean vector $\mu_U$ and covariance matrix $\Sigma_U$. We assume that

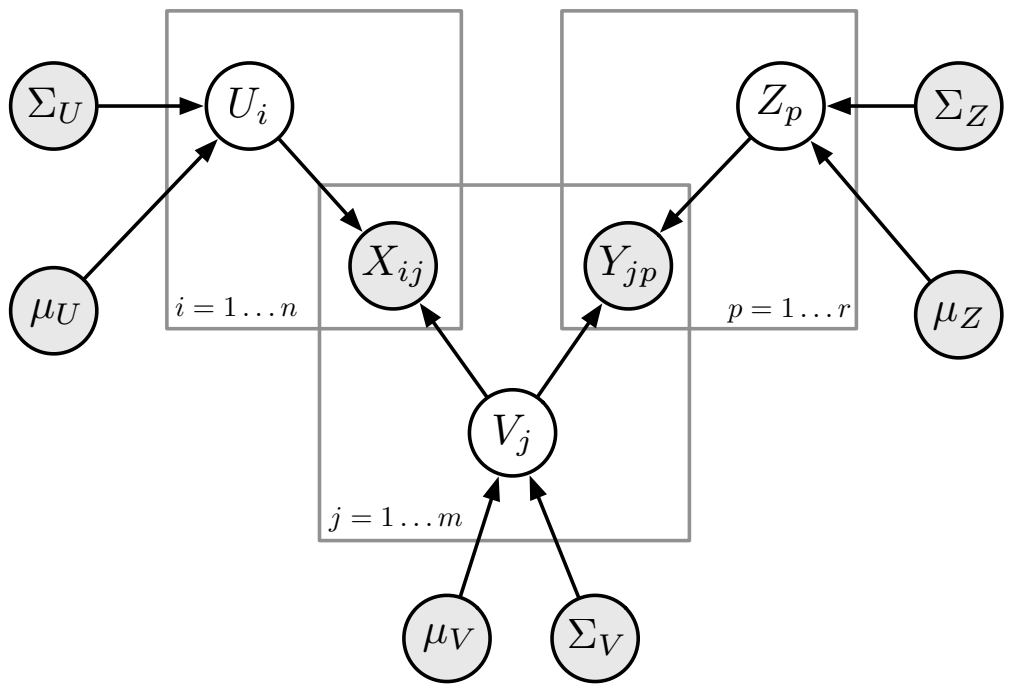

(a) Collective Matrix Factorization

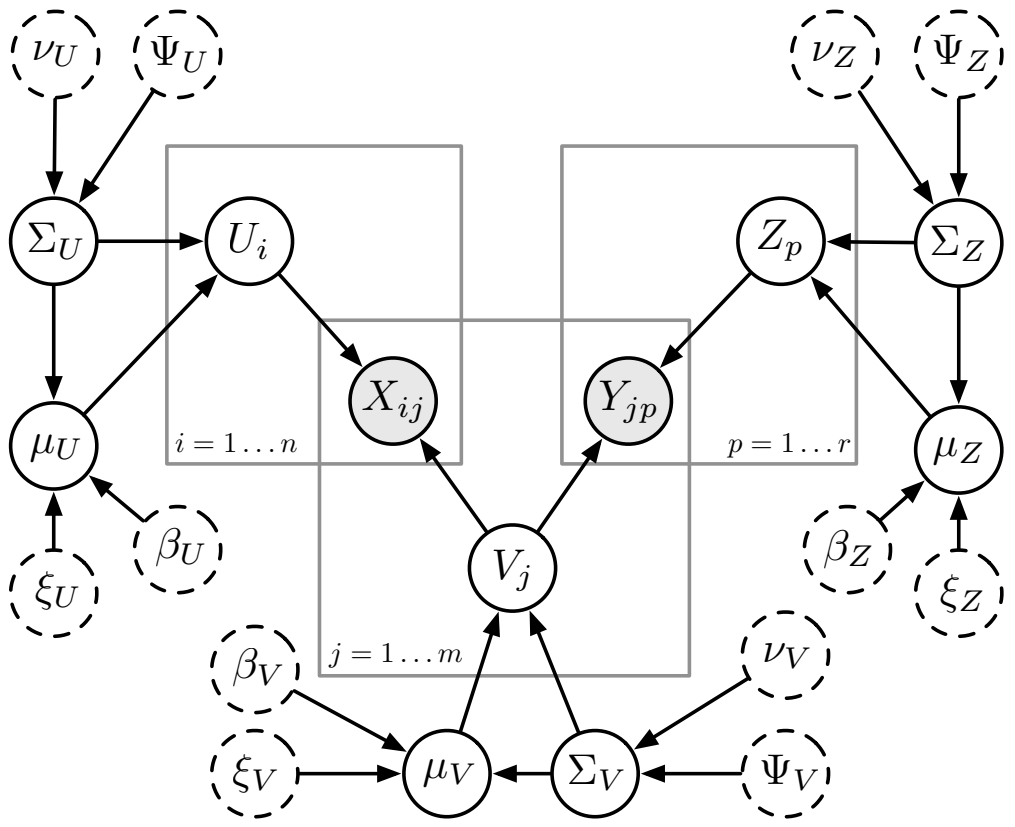

(b) Hierarchical (and Hierarchical Bayesian) Collective Matrix Factorization

Figure 1: Plate graphical models for collective matrix factorization (Section 2) and its analogue with hierarchical priors (Section 3). Shaded nodes indicate known quantities; dashed nodes are fixed parameters in the hyperprior. Weight matrices $W$ and $\tilde{W}$ are elided.

$\Theta_U = (\mu_U, \Sigma_U)$ is known. The priors over $V$ and $Z$ are defined similarly, with $\Theta = (\Theta_U, \Theta_V, \Theta_Z)$. Equations 1–3 define the posterior distribution $\mathcal{P} = p(U, V, Z \mid X, Y, W, \tilde{W}, \Theta)$

$$\begin{aligned} \mathcal{P} &= c \cdot p(X \mid U, V, W)\, p(Y \mid V, Z, \tilde{W}) \\ &\quad p(U \mid \Theta_U)\, p(V \mid \Theta_V)\, p(Z \mid \Theta_Z), \end{aligned} \quad (4)$$

where $c$ is a normalizing constant. Figure 1(a) is a plate representation of Equation 4. Maximum a posteriori inference (a.k.a. regularized maximum likelihood) involves searching for the parameters which minimize the negative log-posterior

$$\mathcal{L} = -\log p(U, V, Z \mid X, Y, W, \tilde{W}, \Theta).$$

The prediction link $f$ is equivalent to the choice of $p_X$: $E[X_{ij}] = f(U_{i\cdot} V_{j\cdot}^T)$ where $f$ is the derivative of the log-partition function of the exponential family of $p_X$ (likewise $g$ and $p_Y$). While $\mathcal{L}$ is non-convex, it is *componentwise convex*: i.e., convex in one low-rank factor when the others are fixed.

Componentwise convexity leads to an elegant algorithm for maximum a posteriori estimation. Consider the graphical model form of collective matrix factorization (Figure 1(a)). Using $d$-separation [12], it is easy to deduce that if only one factor is free (say $U$) then the rows of that factor are independent of one other. Furthermore, the per-row optimization is convex. Therefore, the projection over a large factor matrix can be reduced into parallel convex optimizations over each row of that factor. Each row of a factor has only $k \ll \min\{m, n, r\}$ parameters, and so both the gradient and Hessian may be used to minimize $\mathcal{L}$ with respect to a factor row. We call this approach alternating Newton-projections. In Section 4, the same decomposition into per-row updates will lead to a block Metropolis-Hastings sampler, where the gradient and Hessian computed here are used to select the proposal distribution.

# 3 Hierarchical Collective Matrix Factorization

Collective Matrix Factorization requires choosing a good fixed value for the hyperparameters $\Theta = (\Theta_U, \Theta_V, \Theta_Z)$, where $\forall F \in \mathcal{F}$, $\Theta_F = (\mu_F, \Sigma_F)$. Searching over $\Theta$ is costly, even if the prior means are zero and the covariances spherical.

Another concern is that information between entities can only be shared indirectly, through another factor: e.g., in $f(UV^T)$, two distinct rows of $U$ are correlated only through $V$. Computationally, the independence of rows in the free factor is useful. Statistically, we want a more direct way of pooling shared behaviour among rows or columns of a matrix.

Both concerns can be addressed by extending Collective Matrix Factorization (Figure 1(a)) to include hierarchical priors (Figure 1(b)). Since we do not know $\Theta$, we place a weak prior on it and treat it as a quantity to be learned: i.e., a hierarchical model. The hierarchical prior is designed to preserve independence of factor rows under alternating Newton-projections. We place separate hierarchical priors on $\Theta_U$, $\Theta_V$, and $\Theta_Z$, electing to use the conjugate prior for Gaussian parameters: the normal-Inverse-Wishart distribution [5]. The normal-Inverse-Wishart prior on $\Theta_F = (\mu_F, \Sigma_F)$ is defined by first sampling the covariance from a Wishart distribution, $\mathcal{W}$, then conditionally sampling the mean from a Gaussian distribution, $\mathcal{N}$:

$$\Sigma_F^{-1} \sim \mathcal{W}(\nu_F, \Psi_F),$$
$$\mu_F \,|\, \Sigma_F \sim \mathcal{N}(\xi_F, \Sigma_F/\beta_0).$$

The fixed hyperprior parameters $\nu_F > k$, $\Psi_F \in \mathbb{R}_+^{k\times k}$, $\xi_F \in \mathbb{R}^k$, $\beta_0 > 0$ are chosen by the modeler.[2] Our choice of hierarchical priors for matrix factorization is identical to that of Salakhutdinov and Mnih [16], though we do not make the restrictive simplifying assumption that the likelihood is conjugate to the hierarchical prior.

The hierarchical prior acts as a shrinkage estimator for the rows of a factor, pooling information indirectly, through $\Theta$. Shrinkage may be especially useful when some entities are associated with only a few observations: in the absence of data, the low-rank representation of an entity tends towards the population mean.

Maximum a posteriori estimation of $(\mathcal{F}, \Theta)$ is straightforward: alternate between optimizing $\mathcal{F}$ given fixed $\Theta$, and optimizing $\Theta$ given fixed $\mathcal{F}$. Given fixed $\mathcal{F}$, the most likely value of $(\mu_F, \Sigma_F)$ is the mode of a normal-Inverse-Wishart distribution with parameters

$$\xi_F^* = \frac{n_F}{\beta_F + n_F}\bar{F} + \frac{\beta_F \xi_F}{\beta_F + n_F}, \tag{5}$$

$$\Psi_F^* = \Psi_F^{-1} + S_F + \frac{\beta_F n_F}{\beta_F + n_F}(\bar{F} - \xi_F)^T(\bar{F} - \xi_F), \tag{6}$$

$$\nu_F^* = \nu_F + n_F, \quad \beta_F^* = \beta_F + n_F, \tag{7}$$

$$\bar{F} = \frac{1}{n_F}\sum_{i=1}^{n_F} F_{i\cdot}, \quad S_F = \sum_{i=1}^{n_F}(F_{i\cdot} - \bar{F})^T(F_{i\cdot} - \bar{F}).$$

The fixed hyperprior parameters are referred to as

$$\Theta_0 = (\nu_U, \Psi_U, \xi_U, \nu_V, \Psi_V, \xi_V, \nu_Z, \Psi_Z, \xi_Z, \beta_0).$$

Factoring the posterior distribution over $(\mathcal{F}, \Theta)$, according to the hierarchical model, yields the following objective for maximum a posteriori:

$$\mathcal{O} \;=\; \mathcal{L} + \sum_{F\in\mathcal{F}} \log p(\Theta_F \,|\, \Theta_0).$$

# 4 Bayesian Inference for the Hierarchical Model

Thus far we have discussed only maximum a posteriori estimation, but there may be substantial posterior uncertainty in $(\mathcal{F}, \Theta)$, especially when each entity participates only in a few relationships. The Frequentist argument of asymptotic consistency does not hold for matrix factorization models: the number of parameters grow with the size of $X$ and $Y$.

Moreover, point estimators, like MAP, tend to perform poorly when predicting the behaviour of new entities—here, new rows or columns in the data matrices which did not appear in the training data. Welling et al. [20] provides a compelling theoretical justification for this behaviour.

[2] In all our experiments, for each factor $F$, $\nu_F = k$, i.e., the embedding dimension; $\Psi_F$ is a $k \times k$ identity matrix; $\xi_F = 0$; and $\beta_0 = 1$. The results are not particularly sensitive to the value of the hyperprior parameters.

Finally, point estimation is limited in how it can exploit correlations between the data matrices. Consider Collective Matrix Factorization on the three entity-type example, where the posterior distribution is $p(U, V, Z \mid X, Y, \Theta)$. The analogous posterior involving only the $X$ data matrix is $p(U, V \mid X, \Theta_U, \Theta_V)$. In each case we can compute the marginal distribution of an element of a factor, say $U_{i\ell} \in U$. If we compare the posterior distribution over $U_{i\ell}$ in the single and two-matrix cases, it is clear that the two distributions $p(U_{i\ell} \mid X, Y, \Theta)$ and $p(U_{i\ell} \mid X, \Theta_U, \Theta_V)$ can, and usually will, differ. Under maximum a posteriori inference, the only difference we see between the two distributions is a difference in the mode: changes in the variance, skew, and other properties of the distribution are not accounted for. Changes in the posterior mode account for some of the effect of information sharing between matrices; changes in the full posterior distribution account for all of the effect.

A fully Bayesian approach to training the hierarchical model introduced in Section 3 addresses the aforementioned concerns. Instead of approximating the posterior $p(\mathcal{F}, \Theta \mid \mathcal{D}, \Theta_0)$ by its mode, we approximate it using samples drawn from it.

### 4.1 Block Metropolis-Hastings

In a block Metropolis-Hastings sampler we partition the unknowns $\Omega = (\mathcal{F}, \Theta)$ into subsets (blocks) of correlated variables, cyclically sampling from each block given that the others are fixed. We choose the following blocks for Metropolis-Hastings:

$$\forall F \in \mathcal{F} \;\; \forall i = 1 \ldots n_F : \; F_{i\cdot} \sim p(F_i \mid \Omega - F_{i\cdot}), \tag{8}$$

$$\forall F \in \mathcal{F} : \; \Theta_F \sim p(\Theta_F \mid \Omega - \Theta_F). \tag{9}$$

This grouping of parameters for block Metropolis-Hastings is similar to the grouping in alternating Newton-projections: sample the hyperparameters, then sample each factor by sampling each row of the factor in parallel. Equation 9 is sampling from a normal-Inverse-Wishart; Equation 8 lacks a closed form, unless $X$ and $Y$ are assumed Gaussian—i.e., the data distributions and priors are mutually conjugate. In the single matrix case, with $p_X$ Gaussian, the model is that of Salakhutdinov and Mnih [17].

### 4.2 Hessian Metropolis-Hastings

In our motivating fMRI example, the entries of different data matrices have different response types: binary word co-occurrence, real-valued voxel responses. This flexibility in response type significantly improves predictive accuracy (see Section 6). We do not wish to sacrifice the flexibility of our model, which supports multiple response types, to reap the benefits of Bayesian inference. Instead of assuming that $p_X$ and $p_Y$ are Gaussian, and resorting to Gibbs sampling, we allow $p_X$ and $p_Y$ to be any rank-one exponential family distribution, and consider the more general Metropolis-Hastings sampler.

In Metropolis-Hastings, one often resorts to sampling from a "random walk" proposal distribution: a Gaussian with mean equal to the sample at time $t$, $F_{i\cdot}^{(t)}$ and covariance matrix $v_i \cdot I$. The user must choose $v_i$, for each row, so that the Markov chain mixes quickly. Tuning one proposal distribution is tedious; tuning a proposal distribution for each entity is masochistic. Worse, if the Hessian of the target distribution, with respect to $F_{i\cdot}^{(t)}$, is far from spherical, then the rate at which the underlying Markov chain mixes can be slow, regardless of how $v_i$ is tuned.

The distribution in Equation 8 may not be easy to sample from; but given a point, namely $F_{i\cdot}^{(t)}$, we can easily compute the local gradient and Hessian of the distribution with respect to $F_{i\cdot}$. By using the gradient and Hessian, we can create a proposal distribution that better approximates $p(F_{i\cdot} \mid \Omega - F_{i\cdot})$.

Once one realizes that $p(F_{i\cdot} \mid \Omega - F_{i\cdot})$ is the likelihood of a Bayesian Generalized Linear Model, we can use what we know about efficient inference in Bayesian GLMs to accelerate sampling from Equation 4. In particular, a contribution of this work is the insight that we can use Hessian Metropolis-Hastings (HMH) [14] in Bayesian matrix factorization. HMH uses both the gradient and Hessian to automatically construct a proposal distribution at each sampling step. Intuitively, since we have a efficient Newton-projection for finding the mode of $p(F_{i\cdot} \mid \Omega - F)$, we should use it to pick a proposal that is closer to the mode.

To define a Metropolis-Hastings sampler, we need to define the forward sampling distribution, from which a proposal value $F_{i\cdot}^{(*)}$ is drawn given the previous value in the chain, $F_{i\cdot}^{(t)}$. The choice of a forward sampling distribution leads to a corresponding backward sampling distribution, which defines the probability of returning to $F_{i\cdot}^{(t)}$ from the proposal value $F_{i\cdot}^{(*)}$. The forward sampling distribution in HMH is a Gaussian whose mean is determined by taking one Newton step from $F_{i\cdot}^{(t)}$, and whose covariance is derived from the Hessian used to take the Newton step. The backward sampling distribution in HMH is a Gaussian whose mean is determined by taking one Newton step from $F_{i\cdot}^{(*)}$, and whose covariance is derived from the Hessian used to take the Newton step. Instead of searching over step lengths $\eta$, we sample from a fixed distribution over step lengths (here, uniformly at random). Algorithm 1 describes the block decomposition, which is common to all three approaches. Algorithms 2 and 3 describe the Hessian Metropolis-Hastings sampler for Equation 8.[3]

### 4.3 Generalization to an Arbitrary Number of Relations

Algorithms 1-3 are presented for the general case, which can involve more than two related matrices. Here, we describe the general notation. Entity-types, the different kinds of relation arguments, are indexed by $i = 1 \dots t$. The number of entities of type $i$ in the training set is denoted $n_i$. A matrix corresponding to a relation between entity-types $i$ and $j$ is denoted $X^{(ij)}$. Each relation matrix is represented as the product of low-rank factors:

$$X^{(ij)} \approx f^{(ij)} \left( U^{(i)} \left( U^{(j)} \right)^T \right),$$

where $f^{(ij)}$ is the element-wise link function that maps the low-rank latent representation into predictions. Each factor $U^{(i)}$ has its own Gaussian prior, defined over factor rows. In the hierarchical case, each prior is assigned a normal-Inverse-Wishart hyperprior.

### 4.4 Bayesian Prediction

There are two basic prediction tasks we consider, *hold-out* and *fold-in*. The difference is whether or not the entity being tested was in the training set; fold-in is therefore the more difficult problem.

*Hold-out prediction*: We want to predict the values of relations that an entity participates in, e.g., predict $X_{ij}$. Given a point estimate of the latent factors, the prediction is $\hat{X}_{ij} = f(U_{i\cdot} V_{j\cdot}^T)$. Given the posterior, we integrate out uncertainty:

$$p(X_{ij} \,|\, \mathcal{D}) = \int p(X_{ij} \,|\, \mathcal{F}, \Theta) h(\mathcal{F}, \Theta) \, d\{\mathcal{F}, \Theta\}, \quad (10)$$

$$h(\mathcal{F}, \Theta) = p(\mathcal{F}, \Theta \,|\, X, Y, W, \tilde{W}).$$

Equation 10 is known as the posterior predictive distribution. Since we have only samples from the posterior, $\{(\mathcal{F}^{(s)}, \Theta^{(s)})\}$, we use a Monte Carlo approximation[4],

$$p(X_{ij} \,|\, \mathcal{D}) = \frac{1}{S} \sum_{s=1}^{S} p(X_{ij} \,|\, \mathcal{F}^{(s)}, \Theta^{(s)}).$$

[3] The extended notation used in Algorithms 1–3 are described in Section 4.3.

[4] The number of samples $S$ is $S = 20$ on hold-out experiments, and $S = 10$ on fold-in experiments. The predictive performance did not significantly improve with larger $S$.

**Algorithm 1**: Decomposition Algorithm for MAP and Hessian Metropolis-Hastings

**Input**: Data matrices, $\{X^{(ee')}\}$. The model (embedding dimension $k$, the choice of exponential family for each matrix, and the hyperparameters $\Theta$, if they are fixed).
**Output**: Low rank factors for each entity-type: $U^{(1)} \dots U^{(E)}$.

```
for e = 1...E do
    Initialize U^(e,0) using Algorithm 2 with prior
    mean μ_e = 0 and Σ_e = I. This initialization
    works well for either MAP or Bayesian inference.
while not converged/mixed do
    for e = 1...E do
        foreach row of U^(e,t) : U_i·^(e,t) do
            Update U_i·^(e,t+1) using U_i·^(e,t) and all the
            observations involving the current entity:
            i.e., ∀e', X_i·^(ee'). For MAP, the update is a
            convex optimization, approximated by one
            Newton step; for Bayesian inference, we
            sample from the conditional sampling
            distribution for each factor row
            (Equation 8) using Hessian
            Metropolis-Hastings (Algorithm 3).
        If the hyperparameters for factor U^(e), i.e.,
        (μ_e, Σ_e), are not fixed, then compute the
        normal-Inverse-Wishart posterior with
        parameters defined in Equations 5–7. Use the
        posterior mode for MAP, or a Gibbs sample
        for Bayesian inference.
    t = t + 1
```

*Fold-in prediction*: To generate a low-rank representation for an entity not in the training set under Bayesian inference, we approximate the posterior by fixing the value of $\mathcal{F}^{(s)}$ and $\Theta^{(s)}$ for each sample, and use HMH to generate five samples from the posterior over the new factor row.[5] A fraction of the data involving the new entity is used to generate the posterior samples. Under maximum a posteriori, fold-in reduces to finding the most likely value for a new factor row, given that the learned $(\mathcal{F}, \Theta)$ is fixed.

## 5 Related Work

There is an abundance of work on matrix factorization models; this paper subsumes many of them. The most closely related methods, all based on the bilinear form $X \approx f(UV^T)$, are compared in Table 1.

[5] When sampling the posterior over folded-in entities, we discard the first twenty samples (burn-in), and every other sample after that (subsampling).

**Algorithm 2**: Initialization for Hessian M-H

**Input**: Number of entities of type $e$, $n_e$. Prior mean $\mu_e$ and covariance $\Sigma_e$.

**Output**: Initial value for low-rank factor $U^{(e,0)}$. Mean and negative precision matrix for the first forward sampling distribution: $\bar{U}_{i\cdot}^{(e,0)}$ and $\nabla^2 \mathcal{O}\left(U_{i\cdot}^{(e,0)}\right)$.

**for** $i = 1 \ldots n_e$ **do**

Sample from the prior: $U_{i\cdot}^{(e,0)} \sim \mathcal{N}(\mu_e, \Sigma_e)$.
Choose a random step length: $\eta \sim \mathcal{U}[0,1]$.
Compute gradient and Hessian. Estimate posterior mean using one Newton step:

$$\bar{U}_{i\cdot}^{(e,0)} = U_{i\cdot}^{(e,0)} + \eta \left[\nabla \mathcal{O}\left(U_{i\cdot}^{(e,0)}\right)\right] \left[\nabla^2 \mathcal{O}\left(U_{i\cdot}^{(e,0)}\right)\right]^{-1}.$$

We compare the number of matrices involved (#M), whether the method is Bayesian (Bayes?), the type of prior (Prior/Reg), whether the prior is learned automatically (Hier), whether or not the entries can be modelled using any one-parameter exponential family (Exp), and the optimization techniques used (Optimizer). If the prior is Gaussian, we note whether the covariance is spherical (SC) or diagonal (DC).

The closest competing approach to our Bayesian model is PMDC, which is restricted to modeling entries of a matrix with a Gaussian. The variational Bayesian solution is based on approximating the posterior over each row of a factor as a multivariate Gaussian. Our approach makes a similar Gaussian assumption for the proposal, without having to worry about biases that can be introduced by variational methods.

Our decomposition approach is greatly informed by BMF: we sample the hyperparameters using the same Gibbs step. However, we can model sets of related matrices, without requiring that the predictions be Gaussian. The Hybrid Monte Carlo sampler in BXPCA uses the gradient of the likelihood; we use the gradient and the Hessian for each factor row.

When one of the matrices is a bag-of-words representation, it is common to replace the one-parameter exponential family link with a multinomial link: e.g., pLSI-pHITS [3] ties the parameters of two pLSI models together to add citation information to a pLSI topic model; Latent Dirichlet Allocation (LDA) has been augmented with side information by tying hyperparameters in multiple LDA models [11], or by tying LDA parameters to a matrix factorization [2].

Supervised matrix factorization techniques [13, 15] can be viewed as subsets of the two matrix scenario where one of the matrices consists of labels.

**Algorithm 3**: Hessian Metropolis-Hastings

**Input**: Previous sample from the Markov chain, $U_{i\cdot}^{(e,t)}$. Observations involving entity $i$.

**Output**: Next sample: $U_{i\cdot}^{(e,t+1)}$. Mean and negative precision of the next proposal.

Sample the proposal:

$$U_{i\cdot}^{(e,*)} \sim \mathcal{N}\left(\bar{U}_{i\cdot}^{(e,t)}, \left[-\nabla^2 \mathcal{O}\left(U_{i\cdot}^{(e,t)}\right)\right]^{-1}\right).$$

Compute the gradient $[\nabla \mathcal{O}(U_{i\cdot})]$ and Hessian $\left[\nabla^2 \mathcal{O}(U_{i\cdot})\right]$ at $U_{i\cdot}^{(e,*)}$.

Estimate the posterior mean using one Newton step with random step length, $\eta \sim \mathcal{U}[0,1]$:

$$\bar{U}_{i\cdot}^{(e,*)} = U_{i\cdot}^{(e,*)} + \eta \cdot \left[\nabla \mathcal{O}\left(U_{i\cdot}^{(e,*)}\right)\right] \left[\nabla^2 \mathcal{O}\left(U_{i\cdot}^{(e,*)}\right)\right]^{-1}.$$

Compute the acceptance probability $\alpha = \min\{1, \rho\}$, where $\rho =$

$$\frac{p\left(U_{i\cdot}^{(e,*)} | \mathcal{D}, F^{(t)}, \Theta^{(t)}\right)}{p\left(U_{i\cdot}^{(e,t)} \mid \mathcal{D}, \mathcal{F}^{(t)}, \Theta^{(t)}\right)} \times \frac{\mathcal{N}\left(U_{i\cdot}^{(e,t)} | \bar{U}_{i\cdot}^{(e,*)}, \left[-\nabla^2 \mathcal{O}\left(U_{i\cdot}^{(e,*)}\right)\right]^{-1}\right)}{\mathcal{N}\left(U_{i\cdot}^{(e,*)} | \bar{U}_{i\cdot}^{(e,t)}, \left[-\nabla^2 \mathcal{O}\left(U_{i\cdot}^{(e,t)}\right)\right]^{-1}\right)}$$

**if** $r \sim \mathcal{U}[0,1] \leq \alpha$ **then** $k = *$ **else** $k = t$.

Collect outputs:

$U_{i\cdot}^{(e,t+1)} = U_{i\cdot}^{(e,k)}$, $\bar{U}_{i\cdot}^{(e,t+1)} = \bar{U}_{i\cdot}^{(e,k)}$,
$\nabla^2 \mathcal{O}\left(U_{i\cdot}^{(e,t+1)}\right) = \nabla^2 \mathcal{O}\left(U_{i\cdot}^{(e,k)}\right)$.

## 6 Experiments and Discussion

**Data**: The data collection protocol is described in Mitchell et al. [9]. Stimuli, the shared entities, consist of word-picture pairs flashed onto a screen (e.g., bear, barn, pliers); stimuli are chosen to be exemplars of categories (e.g., animals, buildings, tools). Nine subjects are presented with sixty stimuli. Each subject was presented with each stimulus six times—the fMRI image for a subject given a stimulus is the average of the six presentations. Averaging the voxel response over patients yields the `Response`($stimulus, voxel$) relation. While a fMRI image contains >20,000 voxels, we use only the 500 most stable voxels, following [9]. Training and test voxels are drawn from both hemispheres of the brain, with weak spatial correlation, since voxels associated with vision tend to be more stable.

The `Co-occurs`($word, stimulus$) relation is collected by measuring whether or not the stimulus word occurs within five tokens of a word in the Google Tera-word corpus [1]. Of the ∼50,000 common words in the text corpus, we select 20,000 uniformly at random to reduce the cost of learning. The relations map into two matrices: $X =$ `Co-occurs` and $Y =$ `Response`. Unless stated otherwise, we assume that $X_{ij}$ is Bernoulli distributed, and that $Y_{ij}$ is Gaussian. The embedding dimension is $k = 25$ throughout. We standardize the entries of the $Y$ matrix, to avoid estimating a per-matrix variance as part of the link function $g(\cdot)$.

Table 1: Comparison to related matrix factorization models.

| Algorithm | #M | Bayes? | Prior/Reg | Hier | Exp | Optimizer |
|---|---|---|---|---|---|---|
| PMF [16] | 1 | MLE | Gaussian (SC) | ✓ | ✓ | Gradient descent |
| E-PCA [4] | 1 | MLE | — | ✗ | ✓ | Alt. Bregman |
| BMF [17] | 1 | Bayes | Gaussian | ✓ | ✗ | Gibbs sampling |
| BXPCA [10] | 1 | Bayes | Multiple | ✓ | ✓ | Hybrid MC |
| VBSVD [6] | 1 | Bayes | Gaussian (DC) | ✓ | ✗ | Variational Bayes |
| SoRec [8] | 2 | MLE | Gaussian (SC) | ✗ | ✗ | Gradient descent |
| PMDC [21] | 2+ | Bayes | Gaussian (DC) | ✓ | ✗ | Variational Bayes |
| MRMF [7] | 2+ | MLE | Gaussian (SC) | ✗ | ✗ | Gradient descent |
| CMF [19] | 2+ | MLE | Gaussian (SC) | ✗ | ✓ | Alternating Newton |
| This paper | 2+ | Bayes, MLE | Gaussian | ✓ | ✓ | Alt. HMH/Alt. Newton |

**Evaluation**: In both fold-in and hold-out experiments, our concern is predicting voxel response, i.e., entries of $Y$, under mean squared error. Fold-in experiments involve testing on voxels, columns of $Y$, which did not appear in the training data. In fold-in experiments, two-thirds of a new entity's observations are used in folding-in; the rest for estimating test error. In hold-out experiments, one-tenth of the observations are used to estimate test error.

**Models Compared**: We compare the three approaches discussed in Sections 2–4:

- Hierarchical Bayesian Collective Matrix Factorization (HB-CMF), where we approximate the posterior over $(\mathcal{F}, \Theta)$ using multiple samples from it. [Structure: Figure 1(b); Training: Section 4]
- Hierarchical Collective Matrix Factorization (H-CMF), where we approximate the posterior over $(\mathcal{F}, \Theta)$ using the posterior mode. [Structure: Figure 1(b); Training: Section 3].
- Collective Matrix Factorization (CMF), our baseline model, where we find the most likely value of $\mathcal{F}$ given fixed $\Theta$. [Structure: Figure 1(a); Training: Section 2]. To avoid the computational cost of finding a good value for $\Theta$ by grid search, we use psychic initialization: the $j^{th}$ diagonal element of $\Sigma_F$ is the variance of the $j^{th}$ column of the estimate of $F$ produced by H-CMF. Prior means are fixed to zero.

The only difference between CMF and H-CMF is the hierarchical prior; the only difference between H-CMF and HB-CMF is in how the posterior is approximated.

**Predictive Accuracy**: Figure 2 illustrates how much better the hierarchical Bayesian approach is than point estimate alternatives, on both hold-out and fold-in tasks. Using HB-CMF, a statistically significant improvement is achieved by augmenting the `Response` relation with the `Co-occurs` relation, on both hold-out and fold-in tasks. The results on the fold-in experiment (Figure 2(b)) show that we can achieve high prediction accuracy even when testing voxels that never appeared in the training data.

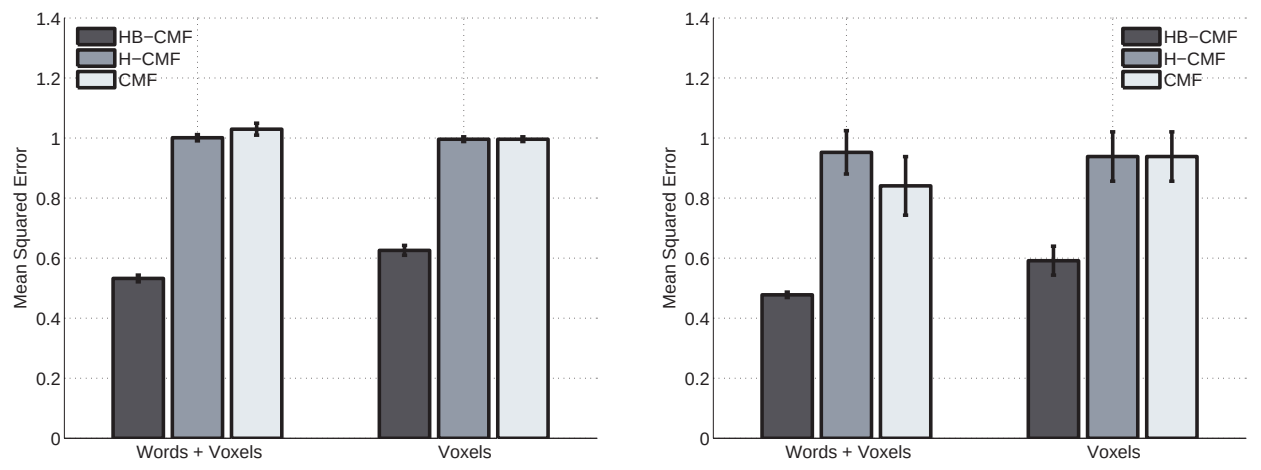

(a) Hold-out (b) Fold-in

Figure 2: Performance on predicting `Response` using just the `Response` relation (Voxel) and augmenting with `Co-occurs` (Words + Voxels). The bars represent algorithms. Error bars are 2-standard deviations, and MSE=1 is chance level.

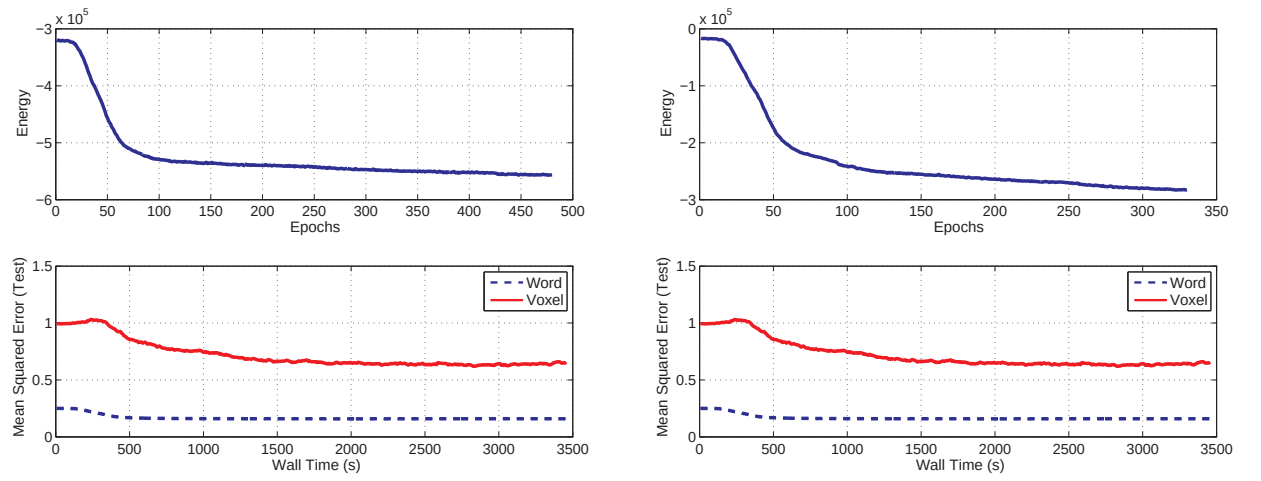

(a) Hold-out (b) Fold-in

Figure 3: Mixing behavior of Hessian M-H. The slowest mixing instance of the hold-out and fold-in experiments are shown. Each point on the energy vs. epochs plots (top) measures the loss of a sample. Each point on the test error vs. time plots (bottom) measures the test error of a sample on predicting `Response` (Voxel) or `Co-occurs` (Word).

The mean over voxel responses (entries in $Y$) is used as a baseline. Since $Y$ has been standardized, the mean predictor is 0, and the mean squared error equals the variance, i.e., 1. Comparing H-CMF and HB-CMF, it becomes clear that on the same model, with the same data, the MAP estimate captures no signal, while the fully Bayesian approach does.

**Importance of Non-Gaussian Response Types**: If one assumed that $X_{ij}$ and $Y_{ij}$ were Gaussian, the argument in favour of the adaptive proposal would be far less compelling. Even if the Gibbs sampling chain did not converge quickly, one could resort to variational Bayes, as in [21]. However, non-Gaussian response types significantly improve predictive accuracy. If, in HB-CMF, we assume that $p_X$ is Gaussian instead of Bernoulli, prediction accuracy decreases: by 26% on hold-out; by 39% on fold-in. While non-Gaussianity complicates the construction of proposal distributions for Metropolis-Hastings, it does have a significant impact on predictive accuracy.

**Computational Cost of Training**: Sampling the parameters and hyperparameters in HB-CMF takes, on average, $7.2s$ for the hold-out experiment using a parallel implementation on four processors.[6] Our implementation was developed in MATLAB, using the Distributed Computing Environment toolkit. An analogous iteration of H-CMF takes $< 10s$. That said, H-CMF converges in less than 20 iterations; HB-CMF takes over 100 iterations to converge. The energy vs. epochs plots (Figure 3) suggest that the underlying Markov chain over states mixes quickly.

## 7 Conclusions

We do not claim a radically new approach for incorporating side information (see Table 1). However, existing methods force one to choose between ignoring parameter uncertainty or making Gaussianity assumptions. In our case study, both uncertainty and non-Gaussianity are critical to prediction; we believe this property is ubiquitous in applications of matrix factorization. This observation motivates our main technical contribution: we address the computational cost of Metropolis-Hastings by constructing a block sampler that exploits both the bilinear structure of matrix factorization and the local structure of the likelihood. A further contribution is the point that, in relational matrix factorization models, true Bayesian inference seems to be necessary to realize the benefit of hierarchy; without our improvements in computational speed, it would have been difficult to test this hypothesis. Finally, in our case study, we have leveraged cheap data to avoid the need for additional, highly expensive fMRI runs; and, we have provided evidence that we can shed light on the brain's internal representation of concepts simply by looking at word co-occurrences.

[6] The four processors are cores on an Opteron 2384 CPU.

## Acknowledgements

We thank T. Mitchell, M. Palatucci, C. Faloutsos, P. Domingos, and Z. Gharahamani for their insightful comments. The authors gratefully acknowledge the support of NSF, under SBE-0836012.

## References

[1] T. Brants and A. Franz. Web 1T 5-gram corpus, version 1. Electronic, Sept. 2006.

[2] J. Chang and D. Blei. Relational topic models for document networks. In *AISTATS*, 2009.

[3] D. Cohn and T. Hofmann. The missing link–a probabilistic model of document content and hypertext connectivity. In *NIPS*, 2000.

[4] M. Collins, S. Dasgupta, and R. E. Schapire. A generalization of principal component analysis to the exponential family. In *NIPS 13*, 2001.

[5] A. Gelman, J. B. Carlin, H. S. Stern, and D. B. Rubin. *Bayesian Data Analysis*. CRC Press, 2nd ed, 2004.

[6] Y. J. Lim and Y. W. Teh. Variational Bayesian approach to movie rating prediction. In *KDD-Cup Workshop*, 2007.

[7] C. Lippert, S. H. Weber, Y. Huang, V. Tresp, M. Schubert, and H.-P. Kriegel. Relation prediction in multi-relational domains using matrix factorization. In *NIPS: Structured Input–Structured Output*, 2008.

[8] H. Ma, H. Yang, M. R. Lyu, and I. King. SoRec: Social recommendation using probabilistic matrix factorization. In *CIKM*, pages 931–940, 2008.

[9] T. M. Mitchell, S. V. Shinkareva, A. Carlson, K.-M. Chang, V. L. Malave, R. A. Mason, and M. A. Just. Predicting human brain activity associated with the meanings of nouns. *Science*, 2008.

[10] S. Mohamed, K. A. Heller, and Z. Ghahramani. Bayesian exponential family PCA. In *NIPS*, 2008.

[11] R. Nallapati, A. Ahmed, E. P. Xing, and W. W. Cohen. Joint latent topic models for text and citations. In *KDD*, pages 542–550, 2008.

[12] J. Pearl. *Probabilistic Reasoning in Intelligent Systems: Networks of Plausible Inference*. 1988.

[13] F. Pereira and G. Gordon. The support vector decomposition machine. In *ICML*, pages 689–696, 2006.

[14] Y. Qi and T. P. Minka. Hessian-based Markov Chain Monte Carlo algorithms. In *First Cape Cod Workshop on Monte Carlo Methods, Cape Cod, Mass.* 2002.

[15] I. Rish, G. Grabarnik, G. Cecchi, F. Pereira, and G. J. Gordon. Closed-form supervised dimensionality reduction with generalized linear models. In *ICML*, pages 832–839, 2008.

[16] R. Salakhutdinov and A. Mnih. Probabilistic matrix factorization. In *NIPS 20*, 2007.

[17] R. Salakhutdinov and A. Mnih. Bayesian probabilistic matrix factorization using MCMC. In *ICML*, 2008.

[18] A. P. Singh. *Efficient Matrix Models for Relational Learning*. PhD thesis, Machine Learning Department, Carnegie Mellon University, 2009.

[19] A. P. Singh and G. J. Gordon. Relational learning via collective matrix factorization. In *KDD*, 2008.

[20] M. Welling, C. Chemudugunta, and N. Sutter. Deterministic latent variable models and their pitfalls. In *SDM*, 2008.

[21] S. Williamson and Z. Ghahramani. Probabilistic models for data combination in recommender systems. In *NIPS: Learning from Multiple Sources*, 2008.